\documentclass[letterpaper,twocolumn,fleqn]{article} 

\usepackage{ist}
\usepackage{times}
\usepackage{amsmath}
\usepackage{amssymb}
\usepackage{url}
\usepackage{subfig}
\usepackage{graphicx}
\usepackage{fixltx2e}
\usepackage{multirow}
\usepackage[export]{adjustbox}
\usepackage[percent]{overpic}

\newcommand{\eg}{{\em e.g.}}
\newcommand{\ie}{{\em i.e.}}

\title{Learning Adaptive Parameter Tuning for Image Processing}

\author{Jingming~Dong$^a$,
        Iuri~Frosio$^b$,
        and~Jan~Kautz$^b$~\\
        \small{$^a$UCLA Vision Lab, University of California, Los Angeles, CA, USA~\\
$^b$NVIDIA, Santa Clara, CA, USA}}

\date{} 

\begin{document} 

\maketitle 

\thispagestyle{empty} 


\begin{abstract}
The non-stationary nature of image characteristics calls for adaptive processing, based on the local image content. We propose a simple and flexible method to learn local tuning of parameters in adaptive image processing: we extract simple local features from an image and learn the relation between these features and the optimal filtering parameters. Learning is performed by optimizing a user defined cost function (any image quality metric) on a training set. We apply our method to three classical problems (denoising, demosaicing and deblurring) and we show the effectiveness of the learned parameter modulation strategies. We also show that these strategies are consistent with theoretical results from the literature.
\end{abstract}

\begin{figure*}
\center
\includegraphics[width=0.9\textwidth, trim = 1.1in 16.7cm 0.7in 2.7cm, clip = true]{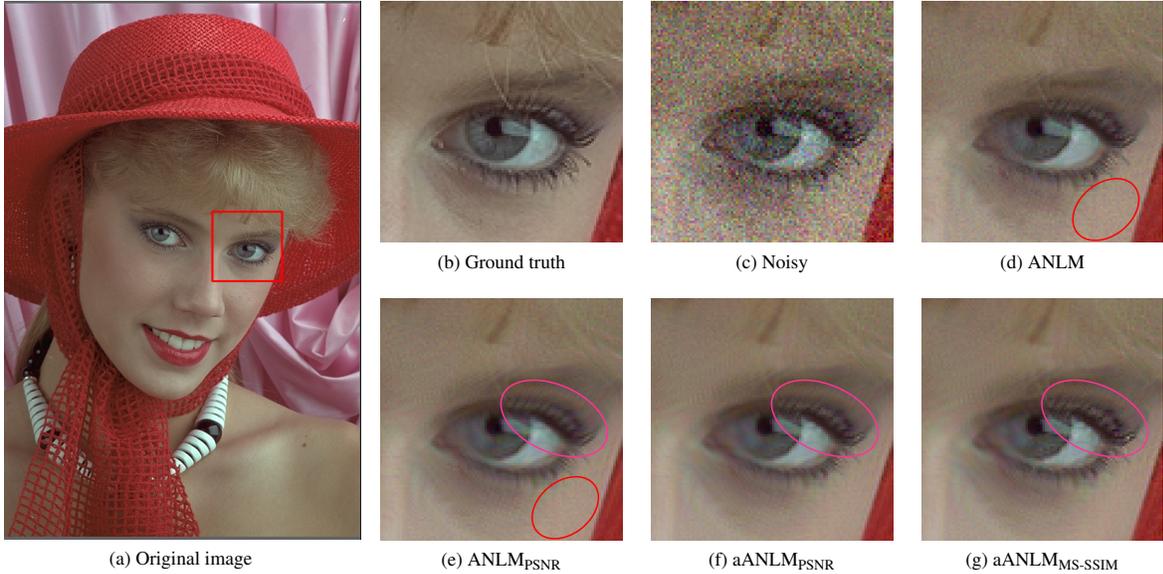}
\caption{Result of Approximate Non Local Mean (ANLM) denoising. ANLM works like standard NLM, but using a limited number (here, 16) of nearest neighbor patches. Panel (d) shows the output of ANLM using the patch size and filtering parameter prescribed in~\cite{Bua05}. Learning the {\em global} patch size and filtering parameter through optimization of PSNR on a training set of images, reduces the noise level on the skin (see the red ellipse for ANLM and ANLM\textsubscript{PSNR}). Learning to {\em adaptively} modulate these parameters increases the contrast of small details (see the pink ellipse on the eyelashes when using {\em globally} learned parameters in (e) ANLM\textsubscript{PSNR}, and {\em adaptively} learned parameters in (f) aANLM\textsubscript{PSNR}). Learning adaptive parameters that optimize for MS-SSIM instead of PSNR further increases the contrast of small details, see (g). Better seen at 4x zoom.}
\label{fig:ANLM}
\end{figure*}

\section{Introduction}
\label{sec:introduction}

The effectiveness of most image processing{\footnote{We refer {\em image processing algorithm} to any algorithm that operates on the image domain, ranging from classical signal processing to any pre-processing step for later computer vision algorithms.}} algorithms depends on a careful parameter choice.
For instance, denoising methods commonly require a \emph{denoising strength} or a \emph{patch size} to be set. These parameters can be adjusted per image, but neglecting the local image characteristics leads to sub-optimal results. Setting the filtering parameters adaptively has obvious benefits; \eg, the denoising strength can be higher in smooth areas where the risk of blurring out details is low, and in turn it can be lower in highly textured areas where noise is less visible. Adaptiveness can also be easily achieved by mixing the output of different algorithms, each operating at best in a different part of an image.

Adaptiveness, however, does not come without a cost. It requires to establish a rule to process the pixels, based on the local image properties. In most cases, this means establishing a mapping from a collection of features, describing the image at a local scale, to a set of parameters that determines the behaviour of the processing algorithm. 
The rule to modulate the parameters across an image can be derived heuristically, or based on a statistical model, but better results are obtained by a learning approach. The learning procedure can in fact be designed to optimize a specific cost function, related to the problem in hand. Nonetheless, previous learning methods have proposed adaptive filtering only for specific image processing problems \cite{Fro15, hammond2007machine, kalantarimachine}.

In contrast, we propose a general approach to learn parameter tuning for adaptive image processing. We learn the mapping from local  features to parameters on a training set, which then generalizes to unseen images. We demonstrate the generality and effectiveness of our approach on three classical image processing problems: denoising, demosaicing and deblurring. We show in the case of denoising that the proposed method is capable of learning a parameter modulation strategy, consistent with previously derived theoretical results~\cite{Duv11}. For demosaicing, we blend the output of three demosaicing methods with adaptive mixing parameters; this does not only produce a better demosaiced image, but also highlights untold strengths and weaknesses in the  demosaicing algorithms. In the case of deblurring, we learn how to adaptively set the regularization hyper-parameter without resorting to any global, discrepancy principle~\cite{Ber10}. Finally, we show how the objective function affects the parameter modulation strategy, coherently with~\cite{Zha15}.

\section{Related work}
\label{sec:relatedWorks}

We mainly identify three approaches\footnote{We are not considering here the class of linear adaptive filters (\eg Wiener or Kalman filters), for which a wide, well established theory exists.} for the development of adaptive algorithms: 1) heuristic-driven, 2) based on local statistics analysis, and 3) learning-based.

\emph{Heuristic-driven $\cdot$}
The rules to modulate the parameters across an image are derived from empirical observation, experience or intuition. This is the case of the adaptive unsharp masking technique in~\cite{Pol00}. The filter was designed to enhance high detail areas, while leaving unaltered the smooth regions. The enhancement strength is guided by the classification of the pixels into low, medium and high dynamic classes. In the denoising context, Chen {\em et al.}\ \cite{Che10} proposed an adaptive Total Variation (TV) regularization method to avoid the staircase artifact introduced by TV. The idea is to identify edges and ramps in an image and apply $\ell_1$ or $\ell_2$ regularization respectively. The blending rule for the two regularization terms is determined heuristically. Although the heuristic approach generally leads to an improvement of the filter performances, it is not suitable to reveal complex or counter-intuitive interactions between the local image characteristics and the filtering parameters~\cite{Fro15}.  

\emph{Analysis of local statistics $\cdot$}
Thaipanich {\em et al.}\ \cite{Tha10} used SVD and K-Means to group similar patches in an image and consequently derived an adaptive Non Local Means (NLM) denoising method. Lebrun {\em et al.}\ described the non-local Bayes denoising algorithm in~\cite{Leb12}, based on the local covariance matrix of a collection of similar patches; they also established the connection between their algorithm and the PCA-based algorithm in~\cite{Zha10}. Duval {\em et al.}\ performed an extended analysis of the filter parameter of NLM denoising, for a simplified patch weight model~\cite{Duv11}. They showed that large patches provide a precise but biased estimate of the noise-free image. The filter parameter has to be consequently modulated across the image to increase the accuracy. To that end, they proposed an iterative method to compute the filter parameter in each pixel. Generally speaking, a careful definition of the local statistics of the signal guarantees superior performances when compared to the heuristic approach, but the price to be paid is the high computational cost of the statistical analysis. Furthermore, complex interactions between the local image characteristics and the optimal filter parameters cannot be identified if not included a-priori in the statistical model. Finally, these algorithms generally optimize a statistical cost function, \eg, $\ell_2$ in ~\cite{Duv11}, that may or may not be related to the application in hand.

\emph{Learning-based $\cdot$}
Using machine learning to learn adaptive processing has the potential to overcome these issues. The main idea is to learn the relation between the image characteristics and the optimal processing procedure at a local scale. Zhang {\em et al.}~\cite{Zha08} used this approach to learn the parameters of an adaptive bilateral filter aiming at denoising and deblurring by minimizing a least squares error over pixel clusters computed from Log of Gaussian responses. Frosio {\em et al.}~\cite{Fro15} showed how to adaptively optimize the spatial and grey level range of the same filter, to maximize the PSNR on a set of noisy training images. The framework described in their paper is applied only to the case of image denoising and Gaussian noise; furthermore, the function putting in relation the local features and the filtering parameters is simpler than the one we propose in Sect.~\ref{sec:method}. The learning approach has also been adopted for Monte Carlo denoising in rendering, where a set of secondary features are extracted from local mean primary features including color, world positions, shading normals among others. These features are mapped to the filter parameters through a perceptron~\cite{kalantarimachine}. Fanello {\em et al.} \cite{Fanello_2014_CVPR} described how to learn optimal linear filters applied to clusters of patches, agglomerated by a random forest. The optimal filters are found by minimizing the $\ell_2$ reconstruction error. Hammond and Simoncelli derived a closed form solution for blending two denoising filters based on wavelets~\cite{hammond2007machine}, whereas Liu {\em et al.}~learned adaptive regularizations for patches based on the estimated distribution of features in transform domain \cite{liu2015image}. It is remarkable to notice that each of the aforementioned approaches is targeted at a unique specific application with a pre-defined metric. On the other hand, our approach is not restricted to a specific application as long as one is able to define a task-specific objective function.

\section{Method}
\label{sec:method}

To process an image, we first extract a set of features describing the characteristics of each pixel in the input image.
We use a parametric model to put in relation the feature vector and the processing parameters for each pixel, and we estimate the model parameters by optimizing a user-defined cost function on a set of training images.

We describe the local image properties in position $(x,y)$ by a feature vector $\mathbf{f}_{x,y} = \left[f^0_{x,y} \ f^1_{x,y} \ ... \ f^{F-1}_{x,y} \right]^\top$ with $F$ elements, where $f^0_{x,y} = 1$ by definition. Typical features are the local image variance and entropies, as in \cite{Fro15}.
The vector $\mathbf{p}_{x,y} = \left[p^0_{x,y} \ p^1_{x,y} \ ... \ p^{P-1}_{x,y} \right]^\top$ contains the $P$ processing parameters for the pixel $\left(x,y\right)$. For instance, in adaptive bilateral filter, the $P = 2$ parameters are the spatial and range sigmas. 
We relate $\mathbf{f}_{x,y}$ and $p^k_{x,y}$ 
with a logistic function:
\begin{equation}
{p}^k_{x,y} = 
{p}^k_{\min}+
\frac{{p}^k_{\max}-{p}^k_{\min}} {1+e^{-h\left({\mathbf{f}_{x,y}}; {\boldsymbol{\theta}_k} \right)}},
\label{eq:logistic}
\end{equation}
and
\begin{equation}
h\left({\mathbf{f}_{x,y}}; {\boldsymbol{\theta}_k} \right) =
{\theta}_k^0 +
{\boldsymbol{\theta}_k^1}^\top \mathbf{f}_{x,y} +
\mathbf{f}_{x,y}^\top {\boldsymbol{\Theta}_k^2} \mathbf{f}_{x,y},
\label{eq:model}
\end{equation}
where $\theta_k^0$ is a scalar, $\boldsymbol{\theta}_k^1$ is a $F \times 1$ vector, $\boldsymbol{\Theta}_k^2$ is a $F \times F$ triangular matrix 
, $\boldsymbol{\theta}_k$ is a vector containing $\theta_k^0$ and the elements of $\boldsymbol{\theta}_k^1$ and $\boldsymbol{\Theta}_k^2$, whereas ${p}^k_{\min}$ and ${p}^k_{\max}$ are the minimum and maximum values of $p^k_{x,y}$ (these are user-defined, reasonable values for a specific filter). We choose the logistic function since it builds a continuous map from $\mathbb{R}^P$ to a bounded interval, while the model in Eq. (\ref{eq:model}) allows representing more complex interactions between features, compared to~\cite{Fro15}.

We denote the set of $M$ input training images with $\{\mathbf{i}^j\}_{j=0}^{M-1}$, while $\{\mathbf{o}^j\}_{j=0}^{M-1}$ is the corresponding set of desired output images. For instance, in the case of denoising, 
$\mathbf{i}^j$ and $\mathbf{o}^j$ are respectively the noisy and noise-free images. Adaptive processing for the $j$-th image is described by:
\begin{equation}
\tilde{\mathbf{o}}^j =
g \left( \mathbf{i}^j; \boldsymbol{\theta}_0, \cdots, \boldsymbol{\theta}_{P-1} \right),
\label{eq:processing}
\end{equation}
where $g$ is the image processing algorithm of interest. 

The learning procedure consists of the optimization of a cost function $E$ with respect to 
$\{ \boldsymbol{\theta}_0, \cdots, \boldsymbol{\theta}_{P-1} \}$
on the set of training pairs 
$\{\left( \mathbf{i}^j, \mathbf{o}^j \right) \}_{j=0}^{M-1}$. 
We typically choose a cost 
$e = E\left(\{\mathbf{\tilde {o}}^j\},\{\mathbf{o}^j\}; \boldsymbol{\theta}_0, \cdots,\boldsymbol{\theta}_{P-1} \right)$ related to the quality of the processed images (for instance, \cite{Fro15} uses the average PSNR of $\{\tilde{\mathbf{o}}^j \}$).
Computing the derivatives of $e$ with respect to the processing parameters may be difficult or even impossible in certain cases: for instance, when the parameters are discrete, as in the case of the patch size for NLM; or when the processing algorithm is iterative, as for TV denoising~\cite{Ber10,Fro13}; or when the cost function is not differentiable, as in the case of FSIM\cite{Zha11b}. Therefore, we resort to a derivative-free optimization algorithm, the Nelder-Mead simplex method \cite{Nel65}. After training, we use the optimal
 $\{ \boldsymbol{\hat \theta}_0, \cdots, \boldsymbol{\hat \theta}_{P-1} \}$
 to compute $\mathbf{p}^k_{x,y}$ for each pixel of any image, out of the training set, through Eqs.~\eqref{eq:logistic} and \eqref{eq:model}. Adaptive filtering is then performed as in Eq.~\eqref{eq:processing}.

\section{Applications}
\label{sec:results}
 
In the following we report results for the proposed learning procedure applied to three classical image processing problems: denoising, demosaicing and deblurring. 


\subsection{Image denoising}

The first application we consider is denoising of color images through NLM~\cite{Bua05}. This filter leverages the image self-similarity to first denoise single patches in the image. Then it averages the denoised patches to get the processed image in a collaborative way. Let's consider a $p^0 \times p^0$ 
patch,~$\mathbf{q}^n$, corrupted by zero mean Gaussian noise with variance $\sigma^2$. The denoised patch, $\mathbf{q}^{d}$, is~\cite{Bua05}:
\begin{eqnarray}
\mathbf{q}^{d} & = &
\sum_{j=1}^N
{w_j~\mathbf{r}_j},\\
\label{eq:NLM}
w_j & = & e^{-{\max \{d^2(\mathbf{q}^n, \mathbf{r}_j)/({p^0})^2 - 2\sigma^2, 0\}}/(p^1\sigma)^2},
\label{eq:NLM_weights}
\end{eqnarray}
where $\{\mathbf{r}_j\}_{j=1}^N$ is a set of $N$ noisy patches, neighbors of $\mathbf{q}^n$, and $d(\mathbf{q}^n, \mathbf{r}_j)$ is the $\ell_2$ distance between $\mathbf{q}^n$ and $\mathbf{r}_j$.
Global (non-adaptive) values for the patch size $p^0$ and the filtering parameter $p^1\sigma$ have been empirically investigated and reported in~\cite{Bua05} as a function of $\sigma$. The choice of these parameters is critical when the Approximate NLM (ANLM) is used, \ie when $N$ is small to reduce the computational cost (\eg, $N = 16$ in~\cite{Tsa15}) and preserve edges better~\cite{Duv11}.

We used the procedure in Sect.~\ref{sec:method} to learn adapting $p^0$ and $p^1$ across an image for ANLM, $N = 16$. The training set was composed of $12$ of the $24$, $512 \times 768$ images in the Kodak dataset~\cite{Pon13}, noise-free and corrupted by zero-mean, Gaussian noise with $\sigma = 20$. We used two different cost functions: the mean PSNR on the training set, $e = \sum_{j=1}^N{\text{PSNR}\left(\tilde{\mathbf{o}}_j, \mathbf{o}_j \right)}$; and the mean MS-SSIM~\cite{Wan03}, $e = \sum_{j=1}^N{\text{MS-SSIM}\left(\tilde{\mathbf{o}}_j, \mathbf{o}_j \right)}$. 
Optimizing for MS-SSIM is particularly interesting in the case of image denoising, as the final consumer of the image is a human observer and MS-SSIM correlates with the human judgement better than PSNR~\cite{Wan03, Zha11b}.
We first performed the optimization using a single, unary feature ($F = 1$), thus not extracting in practice any local feature from the image.
In this case $h \left(\mathbf{f}_{x,y}; \boldsymbol{\theta}_k \right)$ in Eq.~(\ref{eq:model}) is constant and our training procedure boils down to estimating the global, non-adaptive $p^0$ and $p^1$ parameters that maximize PSNR and MS-SSIM on the training set. We refer to these filters respectively as ANLM\textsubscript{PSNR} and ANLM\textsubscript{MS-SSIM}. Then we used $F = 7$ features to describe the local image characteristics. 
These are, apart from the first unary feature, the $3 \times 3$ and $5 \times 5$ local variance, the $3 \times 3$ and $7 \times 7$ entropy and the $3 \times 3$ and $7 \times 7$ gradient entropy. These features are sensitive to edges and textures~\cite{Fro15}. We will refer to these adaptive filters respectively as aANLM\textsubscript{PSNR} and aANLM\textsubscript{MS-SSIM}. Notice that the extracted features are also expected to be noisy in this case. We therefore preprocessed the feature maps with NLM, using a $9 \times 9$ patch and a filtering parameter equal to $0.4 \sigma$, as suggested in~\cite{Bua05}. Since the statistics of the noise in the feature maps is complicated or even unknown, we performed the nearest neighbor search and computed the patch weights using the original, noisy image. 

\begin{table*}
\begin{center}
{\small
\begin{tabular}{|c|c|c|c|c|c|c|}
\hline
& Noisy & ANLM \cite{Bua05} & ANLM\textsubscript{PSNR} & ANLM\textsubscript{MS-SSIM} & aANLM\textsubscript{PSNR} & aANLM\textsubscript{MS-SSIM} \\
\hline
$p^0$ & - & 5 & 9 & 19 & Eq. (\ref{eq:logistic}) & Eq. (\ref{eq:logistic})\\
$p^1 \cdot \sigma$ & - & 0.40$\sigma$ & 0.51$\sigma$ & 0.49$\sigma$ & Eq. (\ref{eq:logistic}) & Eq. (\ref{eq:logistic})\\
\hline
PSNR & 22.11 & 30.06 & 30.27 & 30.00 & \bf{30.80} & 30.48\\
SSIM & 0.6750 & 0.8952 & 0.8992 & 0.9006 & 0.9056 & \bf{0.9084}\\
MS-SSIM & 0.8192 & 0.9417 & 0.9445 & 0.9461 & 0.9480 & \bf{0.9496}\\
\hline
\end{tabular}
}
\caption{Average image quality metrics measured on the testing dataset for ANLM. The values of the patch size $p^0$ and the filtering parameter $p^1$ are reported for the baseline and non-adaptive ANLM. The best result in each row is shown in bold.}
\label{table:ANLM}
\end{center}
\end{table*}

\begin{figure}
\subfloat[aANLM\textsubscript{PSNR}, $p^0$]{
\includegraphics[trim = 2.5cm 1cm 5cm 1cm, clip, width=0.22\textwidth]{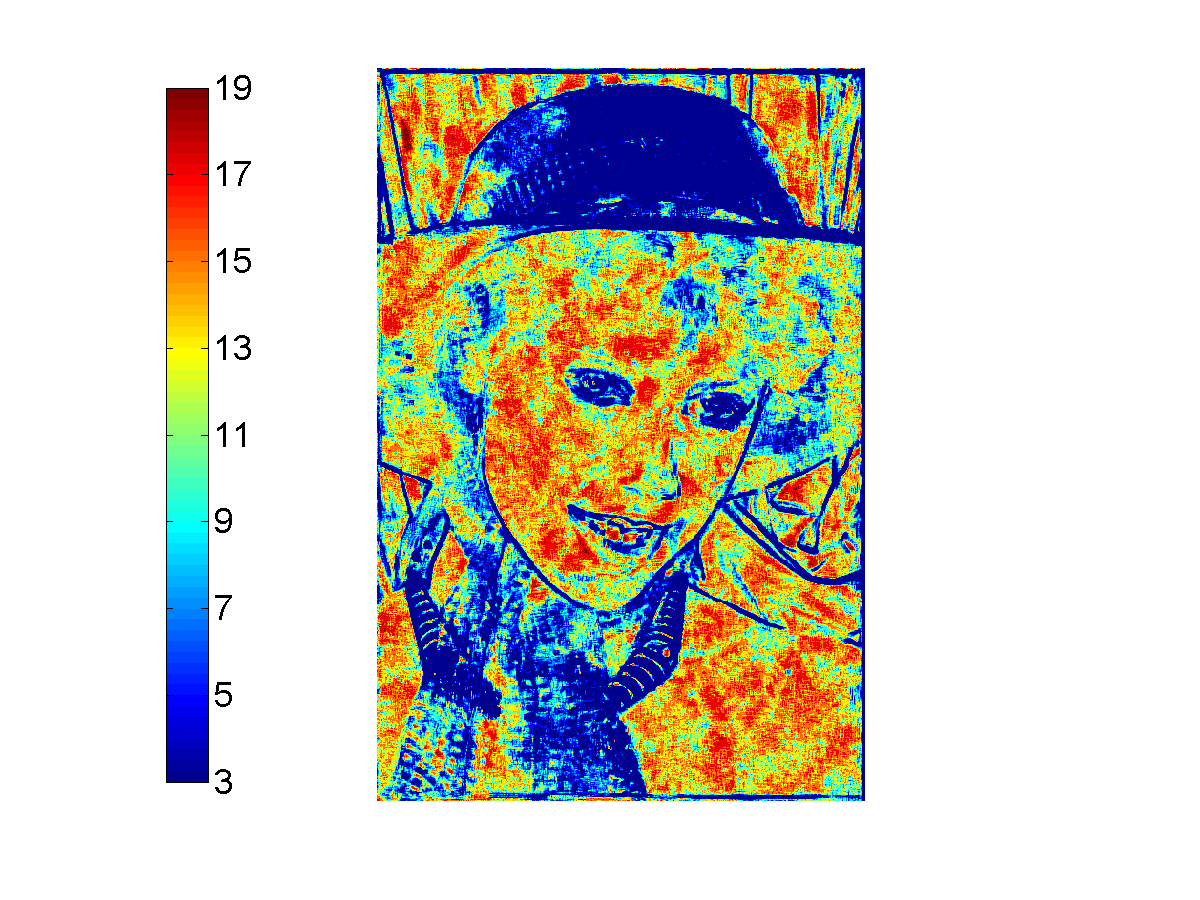}}\rule[0mm]{.1pt}{40mm}%
\subfloat[aANLM\textsubscript{PSNR}, $p^1$]{
\includegraphics[trim = 2.5cm 1cm 5cm 1cm, clip, width=0.22\textwidth]{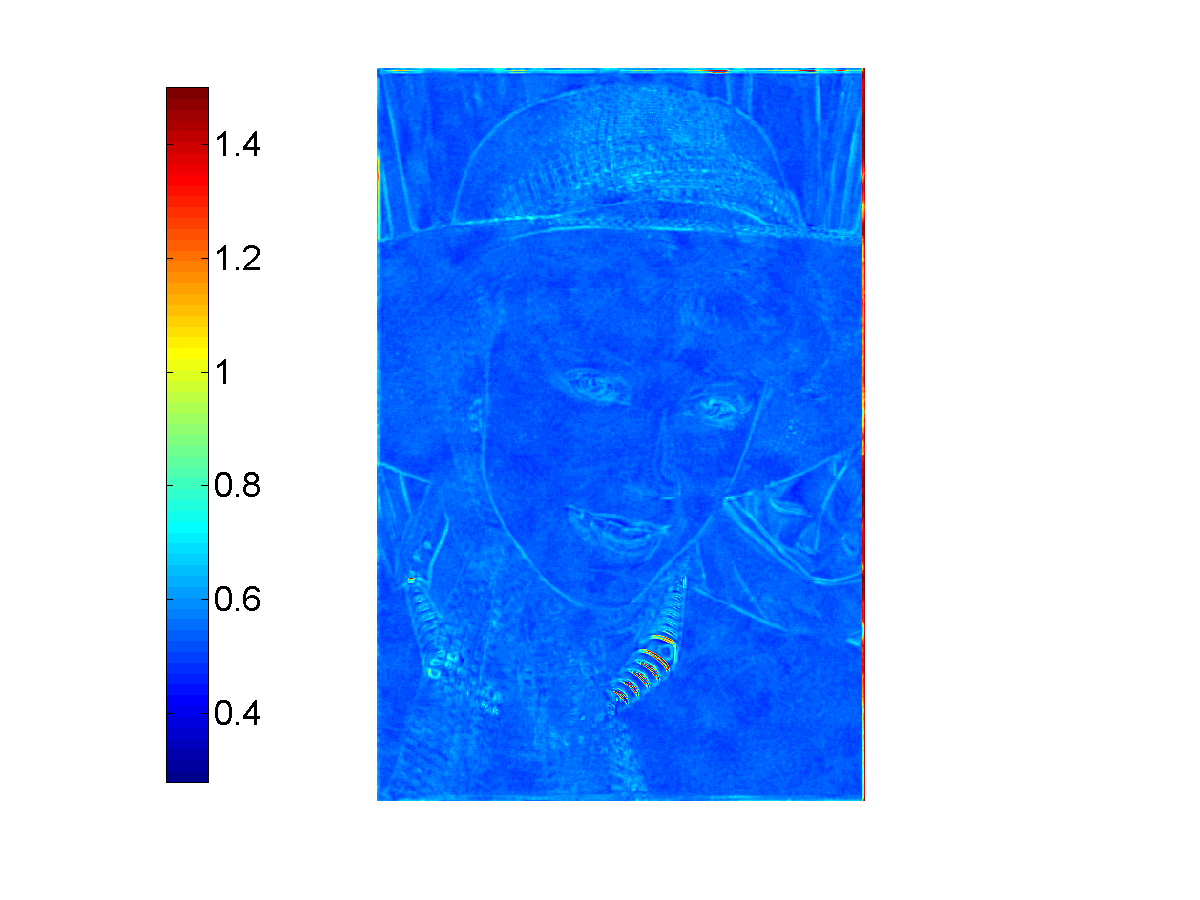}}
\\
\subfloat[aANLM\textsubscript{MS-SSIM}, $p^0$]{
\includegraphics[trim = 2.5cm 1cm 5cm 1cm, clip, width=0.22\textwidth]{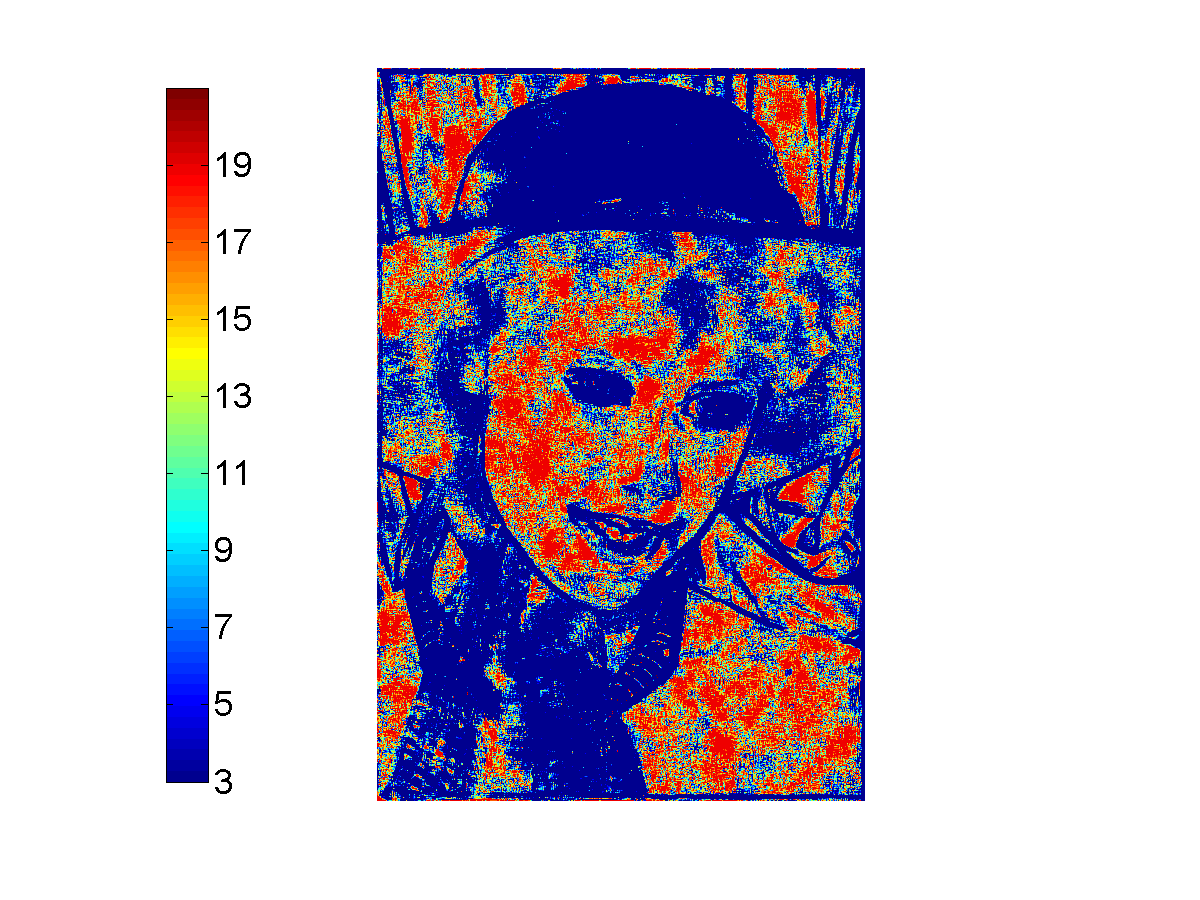}}\rule[0mm]{.1pt}{40mm}%
\subfloat[aANLM\textsubscript{MS-SSIM}, $p^1$]{
\includegraphics[trim = 2.5cm 1cm 5cm 1cm, clip, width=0.22\textwidth]{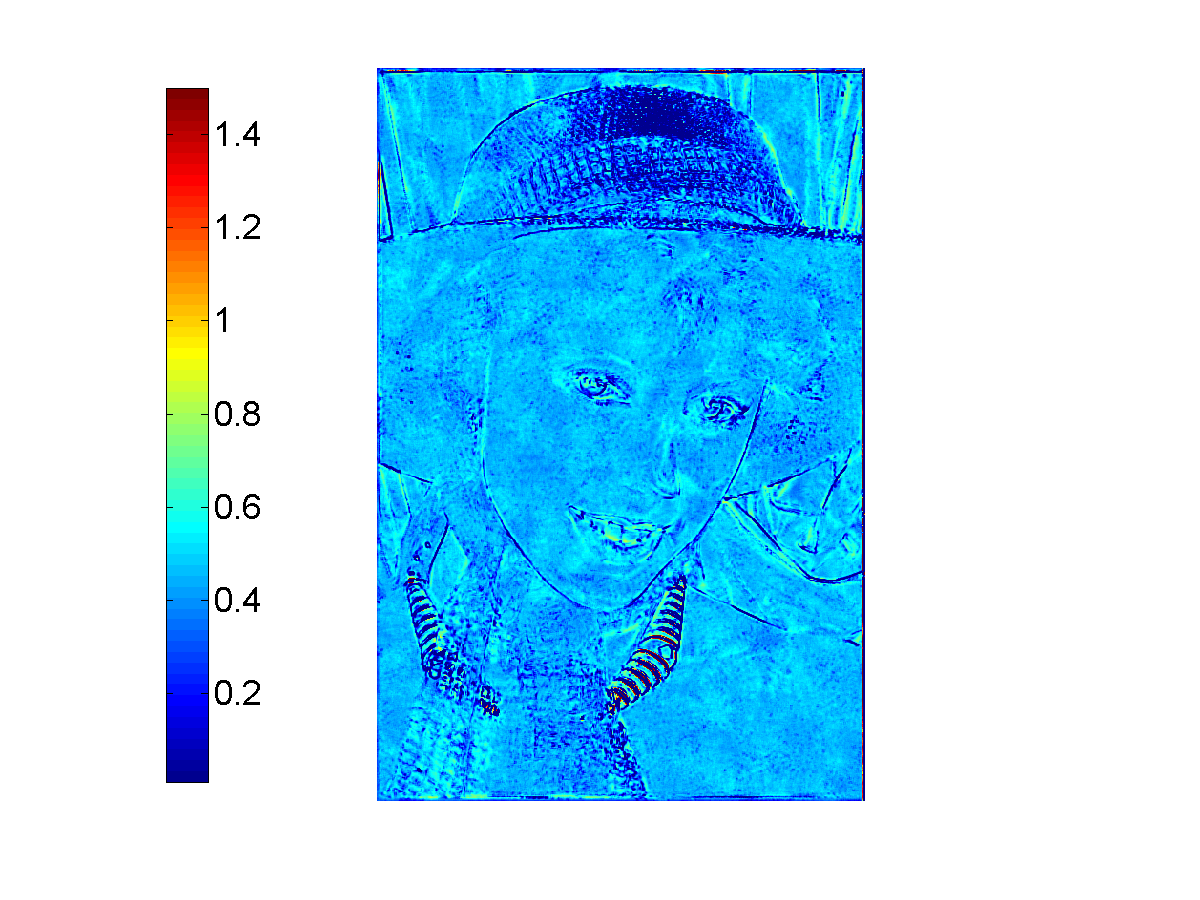}}
\caption{Patch size $p^0$ and filtering parameter $p^1$ learned by aANLM\textsubscript{PSNR} and aANLM\textsubscript{MS-SSIM} for the image in Fig.~\ref{fig:ANLM}.}
\label{fig:ANLMpars}
\end{figure}

We evaluated the performance of ANLM\textsubscript{PSNR}, ANLM\textsubscript{MS-SSIM}, aANLM\textsubscript{PSNR} and aANLM\textsubscript{MS-SSIM} on the $12$ images of the Kodak dataset not used for training. For each filtered image we measured the PSNR, SSIM~\cite{Wan04} and MS-SSIM~\cite{Wan03}. We also compared to ANLM using the patch size $p^0 = 5$ and filtering parameter $p^1 \sigma = 0.4 \sigma$ suggested in~\cite{Bua05}. The average metrics are reported in Table~\ref{table:ANLM}, together with the values of the learned, global (non-adaptive) parameters.
Training a non-adaptive filter to maximize the PSNR leads to an increase in the average PSNR on the testing data, when compared to the original ANLM (from 30.06dB to 30.27dB). This can be visually appreciated in Fig.~\ref{fig:ANLM}, showing that ANLM\textsubscript{PSNR} removes more noise than ANLM in the smooth areas. We observe a similar increase in MS-SSIM for ANLM\textsubscript{MS-SSIM} compared to ANLM (from 0.9417 to 0.9461) but a slight decrease in PSRN (from 30.06dB to 30.00dB).
This is due to the fact that different image quality metrics have different local maxima. In this specific case, optimizing MS-SSIM does not necessarily means optimizing PSNR at the same time.

Turning the global filtering procedure into an adaptive one (from ANLM\textsubscript{PSNR} to  aANLM\textsubscript{PSNR}) results in an additional gain of approximately 0.5dB in PSNR, and an increase in SSIM and MS-SSIM. Fig.~\ref{fig:ANLM} visually demonstrates the advantages of using learning and adaptiveness together: compared to ANLM\textsubscript{PSNR}, aANLM\textsubscript{PSNR} produces slightly more contrasted small details (on the eyelashes in this case). Lastly, Fig. \ref{fig:ANLM}  highlights the differences between optimizing PSNR and MS-SSIM.
MS-SSIM is only partially affected by small residual noise close to edges, which is on the other hand poorly visible by a human observer.
This explains the slightly higher residual noise and, at the same time, the better visual appearance achieved by aANLM\textsubscript{MS-SSIM}, consistent with what was reported in~\cite{Zha15}.

Fig.~\ref{fig:ANLMpars}(a)-(b) shows the parameters learned by aANLM\textsubscript{PSNR} for the image in Fig.~\ref{fig:ANLM}. The learned strategy uses as-large-as-possible patches in the smooth areas (\eg, the skin), and as-small-as-possible close to the edges (\eg, at the face border) and on irregular textures (\eg, the hat). The filtering parameter $p^1$ increases along edges, and it is at maximum in areas where saturation occurs (\eg, on the necklace). This learned strategy is remarkably similar to the one derived by Duval \emph{et al}.~\cite{Duv11}, based on a statistical analysis of NLM. Their main finding is that, close to the edges, we cannot find exact replicas of the reference patch. Consequently, averaging dissimilar patches introduces a bias. The bias can be reduced by using small patches, but this increases the uncertainty because of the poor statistics associated with the smaller number of pixels. To compensate for this, the filtering parameter $p^1$ is increased to include pixels from patches that are far from the reference one. Duval {\em et al}.~\cite{Duv11} demonstrates analytically that such strategy is optimal when a least squares error (or, similarly, PSNR) has to be optimized. Our method is capable of automatically learning the same strategy and even go beyond that. For instance, in the white area of the necklace, the filtering parameter is increased to its maximum since noise is reduced by saturation and the reliability of the measured pixels is consequently higher. 

Even more interestingly, Fig.~\ref{fig:ANLMpars}(c)-(d) shows that a different modulation strategy has to be adopted when a different metric (MS-SSIM in this case) is maximized. Since MS-SSIM is only slightly affected by noise close to edges, the filtering parameter $p^1$ is set to its minimum here, while the patch size $p^0$ is minimal. According to the analysis of Duval {\em et al}., this reduces the bias (\ie, the low pass filtering effect of NLM) and increases the variance (\ie, leave more residual noise) in the filtered image, but it also produces more pleasant images as shown in Fig.~\ref{fig:ANLM}(g). 

\subsection{Image demosaicing}

The second application we consider is the mixture of demosaicing algorithms.
Demosaicing is the process of recovering a full resolution color image from a subsampled (\eg, Bayer) pattern.
Several demosaicing approaches have been proposed in the literature \cite{li2008image}, each with strengths and weaknesses.
Adaptively blending the output of different demosaicing algorithms is expected to achieve a better overall image quality.
For a mixture of $P$ demosaicing algorithms, the output image at pixel $(x, y)$ is defined as: 
\begin{equation}
{\bf \tilde o}_{x,y} = \sum_{k=0}^{P-1} p^k_{x,y}~{\bf d}^k_{x,y} {\large{/}} \sum_{k=0}^{P-1} p^k_{x,y}
\label{eq:demosaicing}
\end{equation}
where ${\bf d}^k_{x,y}$ is the output of the $k$-th algorithm, $p^k_{\min} = 0$ and $p^k_{\max} = 1$, $\forall{k}$.
The learned parameters here are used to compute the blending factors, $p^k_{x,y} {\large{/}} \sum_{k=0}^{P-1} p^k_{x,y}$.

We consider a mixture of $P = 3$ recently published, state-of-the-art demosiacing algorithms: Adaptive Residual Interpolation (ARI,~\cite{Ari15}), an algorithm based on the exploitation  of inter-color correlation (ECC,~\cite{Jai14}), and image demosaicing with contour stencils (CS,~\cite{Get12}).
For training and testing we employed the images of the McMaster dataset~\cite{zhang2011color}, after splitting it into two separate sets.
We first learned how to blend the output of the three algorithms globally (\ie, $F = 1$), by maximizing the mean PSNR (Mix\textsubscript{PSNR}). For adaptive blending, we used as features the local variance, intensity entropy and gradient entropy, computed separately on three channels of the Bayer pattern, for a total of $F = 10$ features.
The local window size for extracting the features is $7 \times 7$, close to the operating scale of most demosaicing algorithms. Learning was performed to maximize the PSNR, SSIM and MS-SSIM on the training data. The corresponding algorithms are indicated as aMix\textsubscript{PSNR}, aMix\textsubscript{SSIM}, and aMix\textsubscript{MS-SSIM}. Results are reported in Table \ref{tbl-mcm}.

The learned, adaptive mixture, aMix\textsubscript{MS-SSIM}, outperforms the best of the three original methods by  $0.96$dB on average in terms of PSNR, when MS-SSIM is maximized in training.
The improvement is consistent for all the three metrics considered here (PSNR, SSIM and MS-SSIM) even if only one of them is optimized during training.
Consistent with the case of denoising, training to maximize SSIM or MS-SSIM produces the best results in terms of SSIM or MS-SSIM.
This also produces an improvement in terms of PSNR since these image quality metrics are correlated, even if they do not share the same local maxima.
Remarkably, even without any adaptiveness, the learned mix (Mix\textsubscript{PSNR}) gains $0.86$dB on average in terms of PSNR, compared to the best of ECC, ARI and CS. The  learned (non-adaptive) blending factors are in this case 0.50 for ECC, 0.44 for ARI and 0.06 for CS. The low importance given to CS on the average is somehow expected, given the lower performance achieved by this method and reported in Table \ref{tbl-mcm}. On the other hand, the fact that the blending factor for CS is non-zero demonstrates that CS can be coupled with the other two algorithms to increase the image quality. This advantage is maximized when turning to an adaptive approach, which results into an additional $0.1$dB on the average compared to the non adaptive approach. 

\begin{figure}[t]
\centering
\begin{tabular}{cc}
Image & Blending factors\\
& (ECC=r, ARI=g, CS=b)\\
{\includegraphics[width=.46\columnwidth]{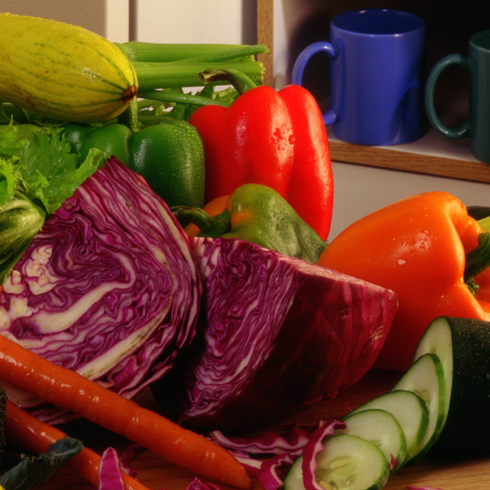}}&
{\includegraphics[width=.46\columnwidth]{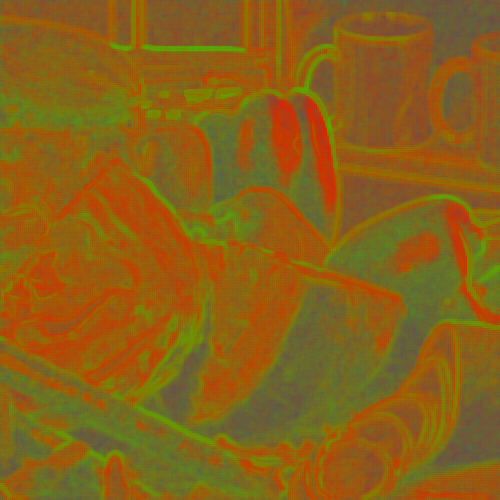}}\\
{\includegraphics[width=.46\columnwidth]{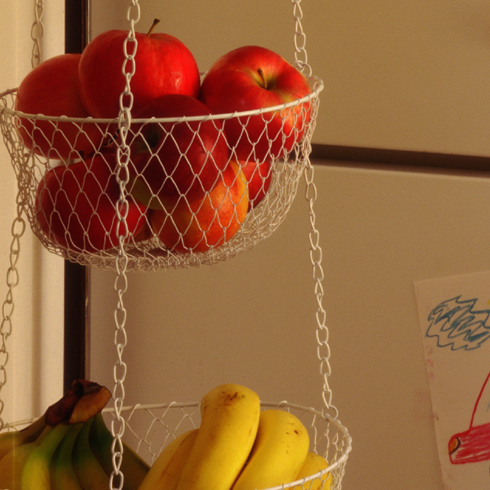}}&
{\includegraphics[width=.46\columnwidth]{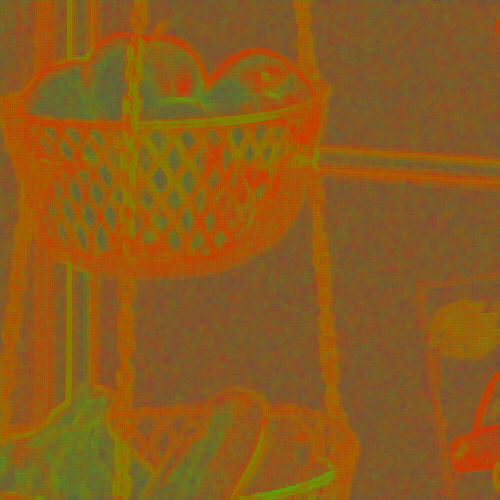}}\\
{\includegraphics[width=.46\columnwidth]{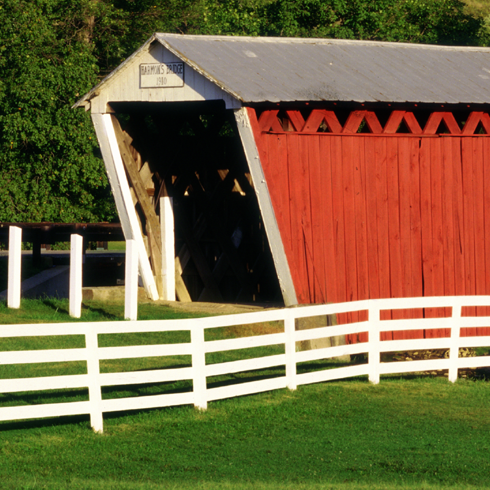}}&
{\includegraphics[width=.46\columnwidth]{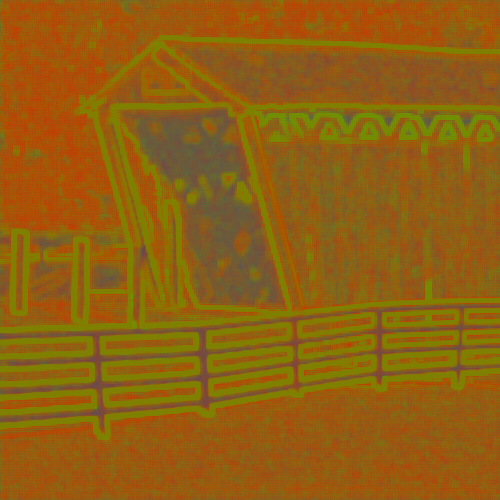}}\\
\end{tabular}
\caption{Blending factors learned by aMIX\textsubscript{MS-SSIM} (right column) for three images of the McMaster dataset~\cite{zhang2011color}. The blending factor for ECC, $p^0 / \sum{p^k}$, is associated with the red channel, whereas those of ARI ($p^1 / \sum{p^k}$) and CS ($p^2 / \sum{p^k}$) are associated with the green and blue channels, respectively. ECC and ARI are the preferred methods for edges and textured areas, whereas CS has more importance in very dark or bright areas of the images.}
\label{fig-params}
\end{figure}

\begin{figure*}[th!]
\centering
\setlength{\tabcolsep}{0.07em}
\begin{tabular}{ccc}
Image & Ground truth & ECC\\
\includegraphics[width=0.25\textwidth]{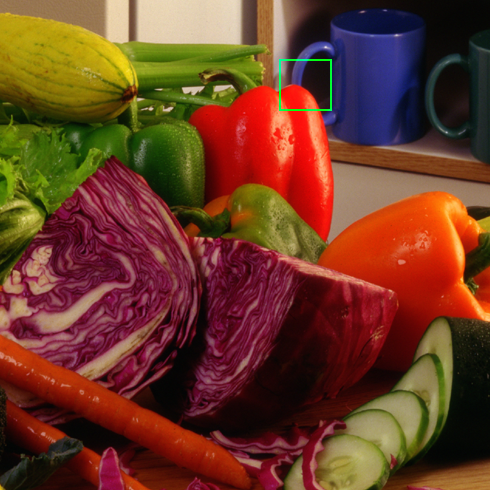}&
\includegraphics[width=0.25\textwidth]{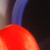}&
\begin{overpic}[width=0.25\textwidth]{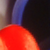}
\put (50,50){\includegraphics[width=0.12\textwidth]{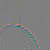}}
\end{overpic}
\\
ARI & CS & Mix\textsubscript{PSNR}\\
\begin{overpic}[width=0.25\textwidth]{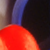}
\put (50,50){\includegraphics[width=0.12\textwidth]{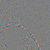}}
\end{overpic}&
\begin{overpic}[width=0.25\textwidth]{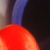}
\put (50,50){\includegraphics[width=0.12\textwidth]{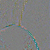}}
\end{overpic}&
\begin{overpic}[width=0.25\textwidth]{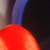}
\put (50,50){\includegraphics[width=0.12\textwidth]{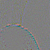}}
\end{overpic}\\
aMix\textsubscript{PSNR} & aMix\textsubscript{SSIM} & aMix\textsubscript{MS-SSIM}\\
\begin{overpic}[width=0.25\textwidth]{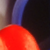}
\put (50,50){\includegraphics[width=0.12\textwidth]{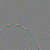}}
\end{overpic}&
\begin{overpic}[width=0.25\textwidth]{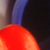}
\put (50,50){\includegraphics[width=0.12\textwidth]{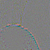}}
\end{overpic}&
\begin{overpic}[width=0.25\textwidth]{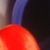}
\put (50,50){\includegraphics[width=0.12\textwidth]{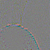}}
\end{overpic}\\
\end{tabular}
\caption{Output of different demosaicing algorithms for the green patch in the upper left panel. The inset panels shows the residual error. The learned blending strategy gains approximately 1dB on the average compared to previous state-of-the-art results. Better seen at $4\times$ zoom.}
\label{fig-mcm}
\end{figure*}

Fig.~\ref{fig-params} shows the blending factors for three testing images. ECC and ARI are the best methods when used alone, and they are almost equally distributed in the mixture. Close to the edges, the learned modulation strategy select either one or the other of these methods. In homogeneus areas they appear to be equally important, with the exception of very dark or bright areas, where the weights associated to CS becomes predominant, even if this is the worst algorithm (among the three considered here) when used alone. The quality of the images obtained with the proposed method can be appreciated in Fig. \ref{fig-mcm}, showing a better reconstruction of the border of the red pepper compared to state-of-the-art ECC and ARI.

\begin{table*}
\begin{center}
{\small
\begin{tabular}{|c|c|c|c|c|c|c|c|}
\hline
& ECC~\cite{Jai14} & ARI~\cite{Ari15} & CS~\cite{Get12} & Mix\textsubscript{PSNR} & aMix\textsubscript{PSNR} & aMix\textsubscript{SSIM} & aMix\textsubscript{MS-SSIM}\\
\hline
PSNR & 38.85 & 38.37 & 36.67 & 39.71 & \bf{39.81} & 39.80 & \bf{39.81}\\
SSIM & 0.9666 & 0.9633 & 0.9516 & 0.9717 & \bf{0.9725} & \bf{0.9725} & \bf{0.9725}\\
MS-SSIM & 0.9951 & 0.9945 & 0.9913 & 0.9963 & 0.9963 & 0.9963 & \bf{0.9964}\\
\hline
\end{tabular}
}
\caption{Average image quality metrics on the testing dataset for demosaicing. The best result in each row is shown in bold.}
\label{tbl-mcm}
\end{center}
\end{table*}

\subsection{Image deblurring}

The last application we consider is image deblurring through TV regularization. This problem is of particular importance in the medical and astronomical fields, where imaging apparatuses with a known point spread function measure single channel images with a very limited number of photons~\cite{Ber10,Fro13}. The problem can be stated as follow. Given the vectorial representation of a noise-free image, $\bf{o}$, the measured, noisy, and blurred image, $\bf{i}$, is:
\begin{equation}
\mathbf{i} = \bf{H} \cdot \mathbf{o} + \mathbf{n}
\label{eq:deblurring}
\end{equation}
where $\mathbf{H}$ is a matrix representing a linear blur and $\mathbf{i}$ is corrupted by photon counting noise $\mathbf{n}$\footnote{With an abuse of notation, we indicate here the photon counting noise as additional noise $\mathbf{n}$.} 
with a Poisson distribution~\cite{Ber10}. The inverse problem of estimating $\mathbf{o}$ from $\mathbf{i}$ is formulated in a Bayesian context as a \emph{maximum a posteriori} problem. We estimate the noise free image, $\tilde{\mathbf{o}}$, as:
\begin{equation}
\tilde{\mathbf{o}} = \arg \min_{\hat{\mathbf{o}}} -\log(L \left(\mathbf{i}|\hat{\mathbf{o}} \right)) + p_{x,y}^{0} R \left( \hat{\mathbf{o}} \right),
\label{eq:deblurringCostFunction}
\end{equation}
where $L \left(\mathbf{i}|\hat{\mathbf{o}} \right)$ is the likelihood term (the Kullback-Leibler divergence for Poisson noise) and $R \left( \hat{\mathbf{o}} \right)$ is the regularization term.
TV regularization is often adopted for its edge preserving capability.
The regularization hyper-parameter $ p_{x,y}^{0}$ is traditionally kept constant across the image and chosen, for instance, through a discrepancy principle~\cite{Ber10}. For simplicity, we solve  Eq.~\eqref{eq:deblurringCostFunction} by steepest descent, even if more efficient algorithms can be used. 

We used the procedure from Sect.~\ref{sec:method} to learn how to adapt $p^0_{x,y}$ across the image.
The training dataset was composed of $12$ of the $24$ images of the Kodak dataset, converted to grey levels.
Each image was blurred with a Gaussian kernel of $7 \times 7$ pixels and a standard deviation of $2$ pixels.
White, photon counting noise was added to the blurred images, assuming a maximum number of $1024$ photons per pixel.
During learning we optimized the average PSNR and MS-SSIM on the training dataset. We first performed training for $F = 1$ (\ie for a global, non-adaptive hyper-regularization parameter $p^0$).
We refer to these filters as TV\textsubscript{PSNR} and TV\textsubscript{MS-SSIM}.
We then performed training using $F = 7$ features: the unary feature and the $5 \times 5$ and $9 \times 9$ local grey level average, standard deviation and their ratio.
The rationale for these features is that the optimal regularization hyper-parameter may be affected by the signal-to-noise ratio (which is proportional to the number of photons) and by the presence of structure in the local image (associated with the local variance).
We refer to the corresponding adaptive filters as aTV\textsubscript{PSNR} and aTV\textsubscript{MS-SSIM}.

\begin{figure*}[th]
\renewcommand{\arraystretch}{0}
\begin{adjustbox}{valign=t}
\begin{tabular}{ccc}
\subfloat[Original image]{\includegraphics[height=82pt]{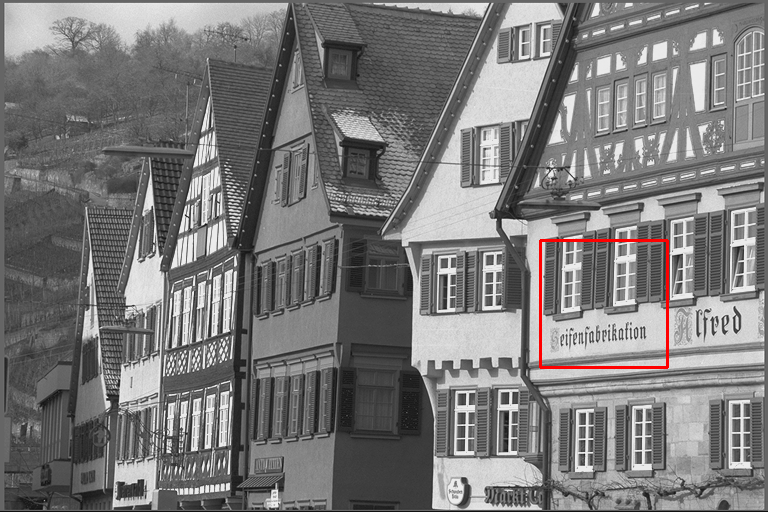}} &
\subfloat[aTV\textsubscript{PSNR}, $p^0$.]{\includegraphics[trim = 1.8cm 3cm 3cm 2.5cm, clip, height=82pt]{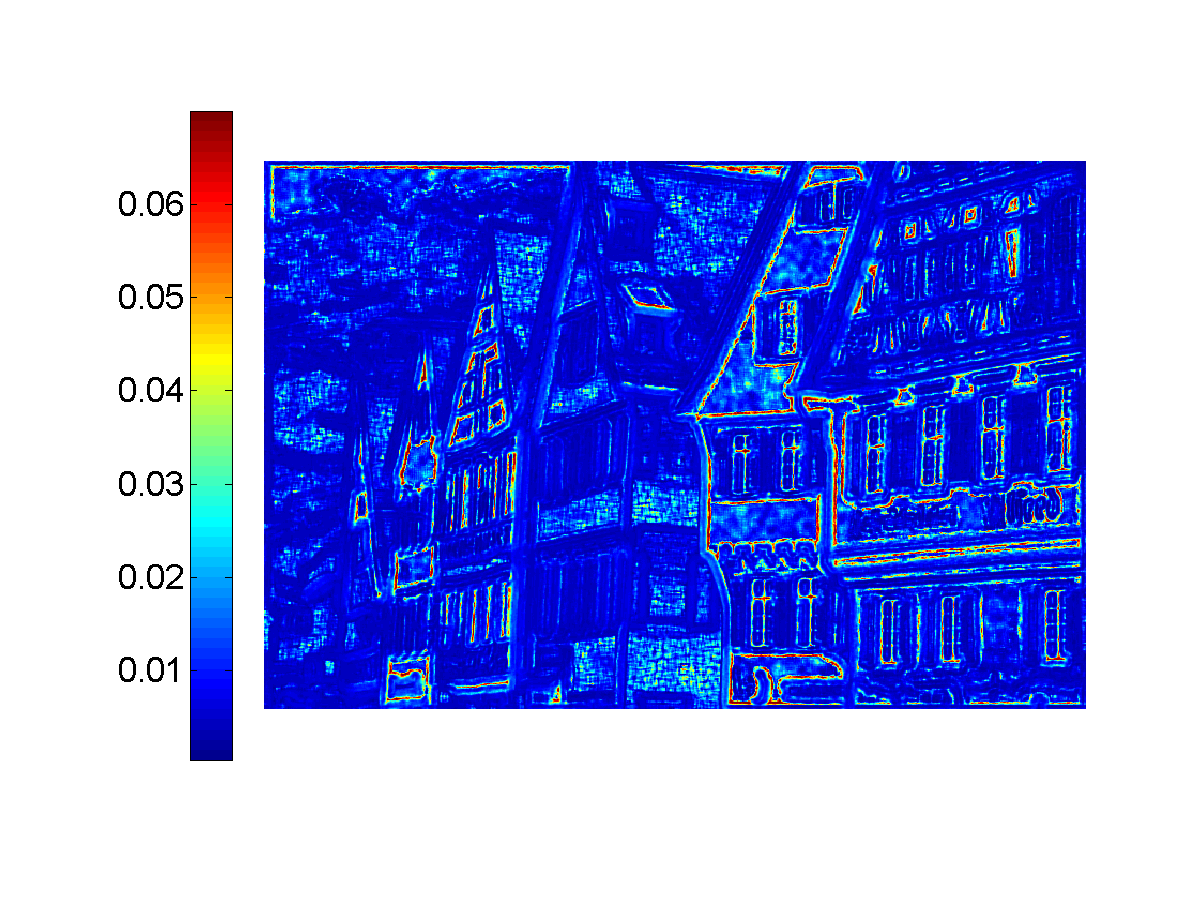}}&
\subfloat[aTV\textsubscript{MS-SSIM}, $p^0$.]{\includegraphics[trim = 1.8cm 3cm 3cm 2.5cm, clip, height=82pt]{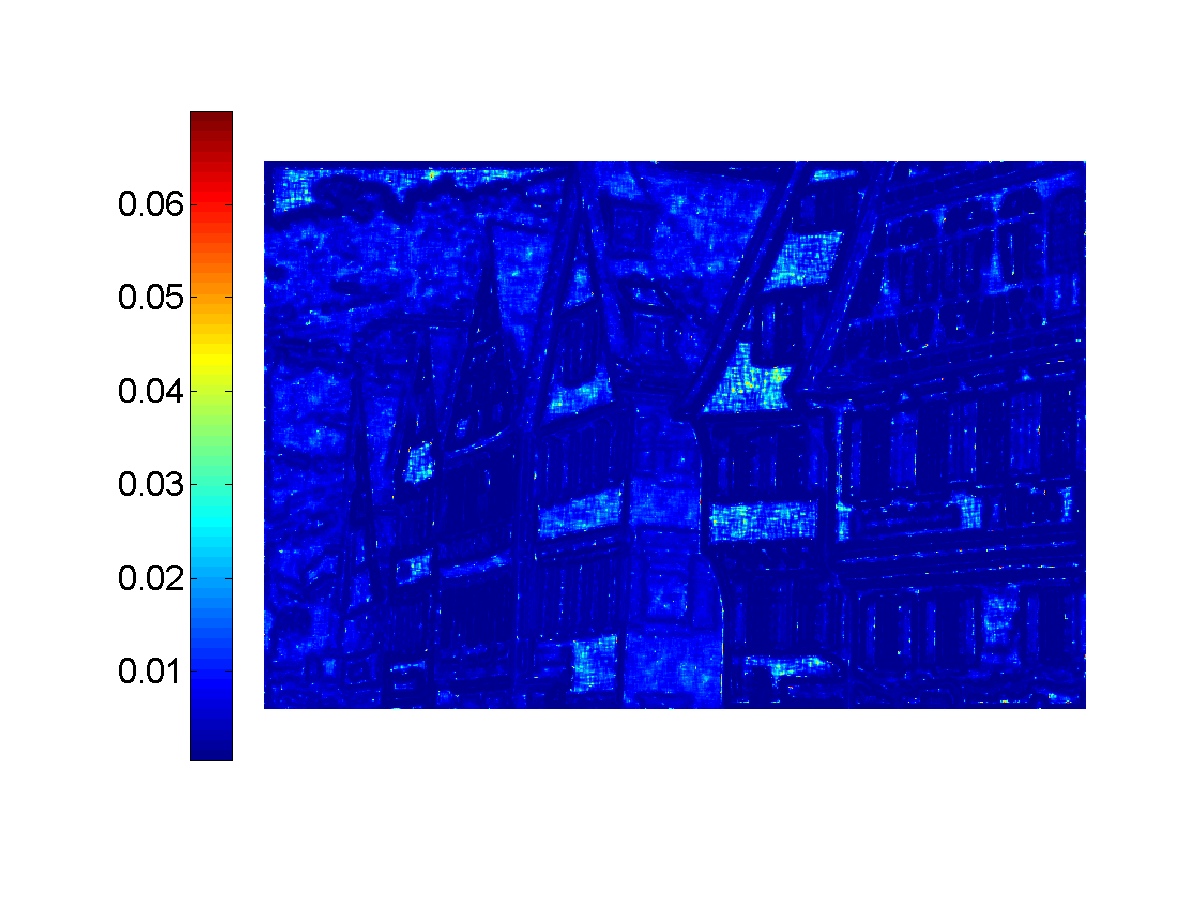}}\\
\begin{tabular}{cc}
\subfloat[Ground truth]{\includegraphics[height = 62pt]{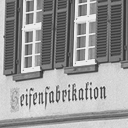}}&
\subfloat[Noisy, blurred]{\includegraphics[height = 62pt]{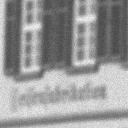}}
\end{tabular}
&
\begin{tabular}{cc}
\subfloat[TV\textsubscript{PSNR}]{\includegraphics[height = 62pt]{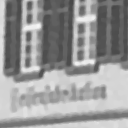}} &
\subfloat[TV\textsubscript{MS-SSIM}]{\includegraphics[height = 62pt]{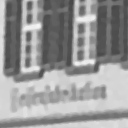}}
\end{tabular}
&
\begin{tabular}{cc}
\subfloat[aTV\textsubscript{PSNR}]{\includegraphics[height = 62pt]{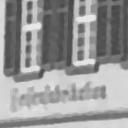}}&
\subfloat[aTV\textsubscript{MS-SSIM}]{\includegraphics[height = 62pt]{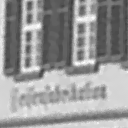}}
\end{tabular}\\
\subfloat[Original image]{\includegraphics[height=82pt]{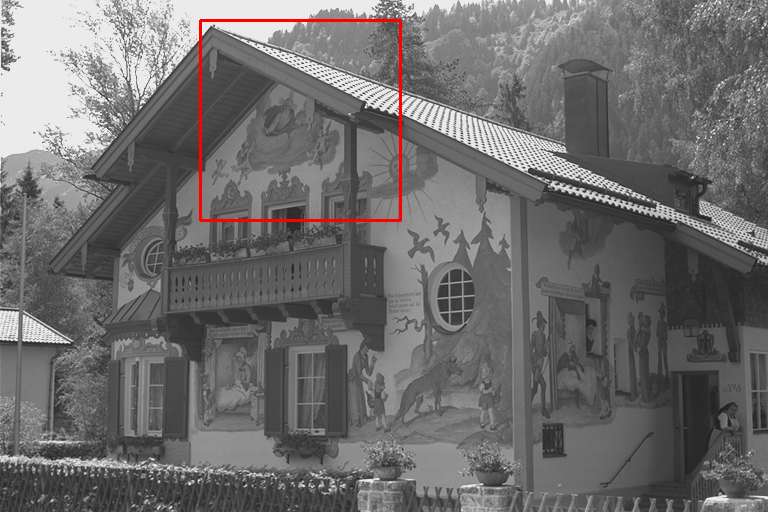}} &
\subfloat[aTV\textsubscript{PSNR}, $p^0$.]{\includegraphics[trim = 1.8cm 3cm 3cm 2.5cm, clip, height=82pt]{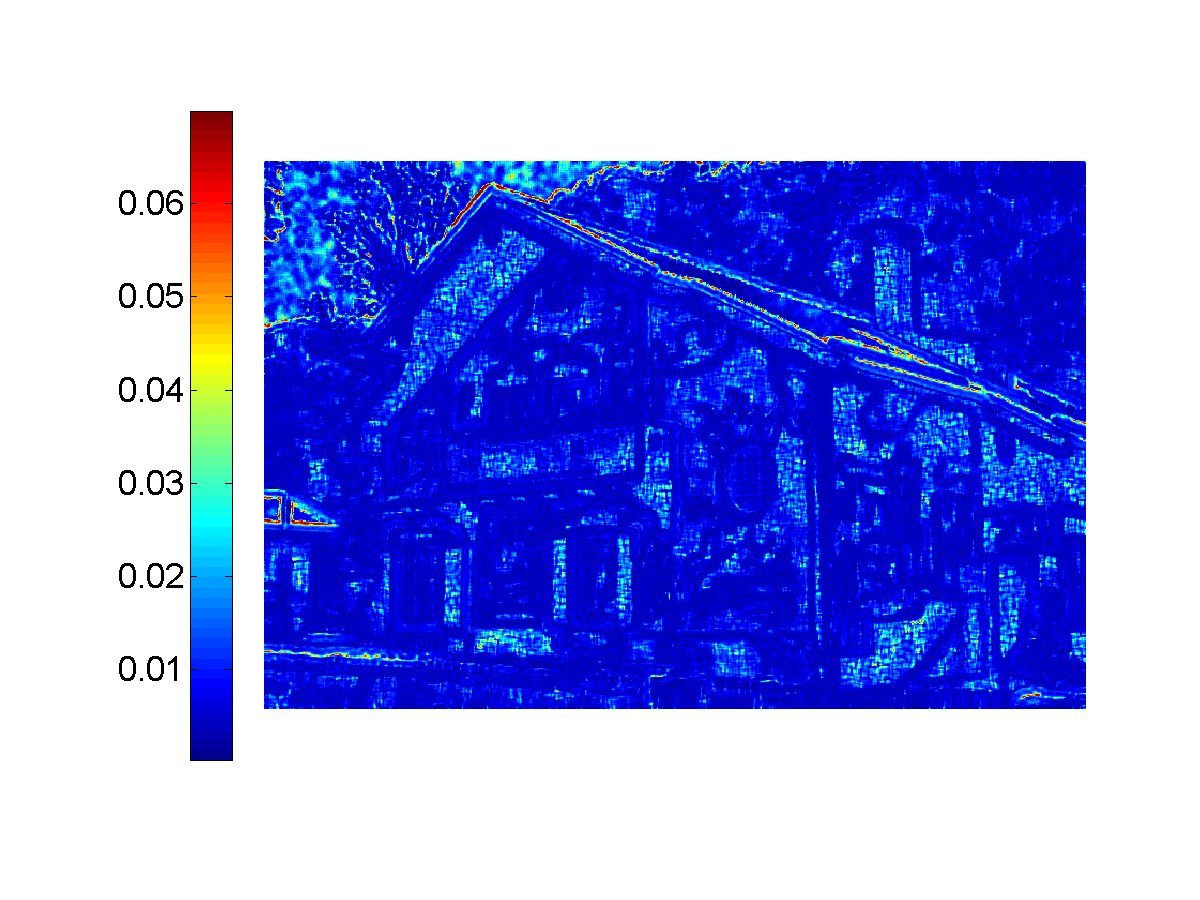}}&
\subfloat[aTV\textsubscript{MS-SSIM}, $p^0$.]{\includegraphics[trim = 1.8cm 3cm 3cm 2.5cm, clip, height=82pt]{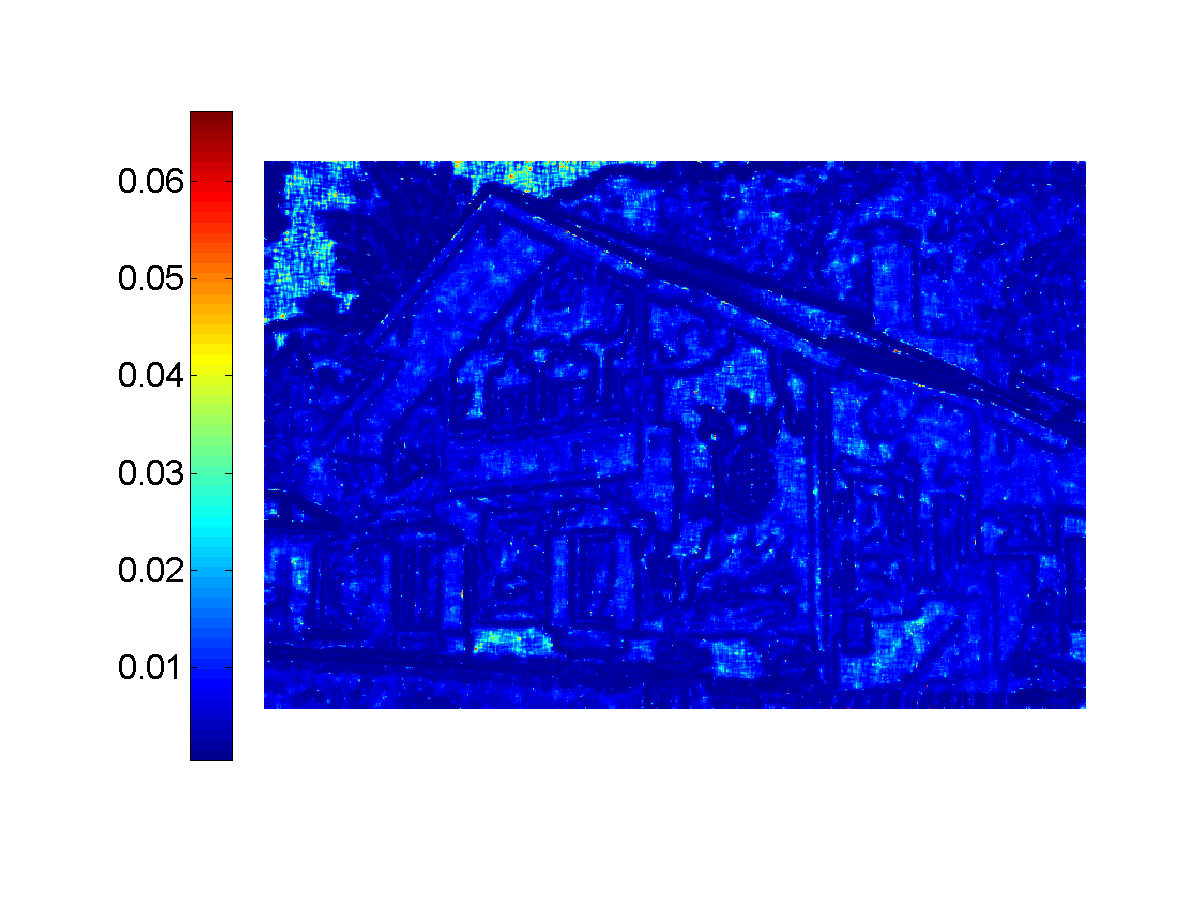}}\\
\begin{tabular}{cc}
\subfloat[Ground truth]{\includegraphics[height = 62pt]{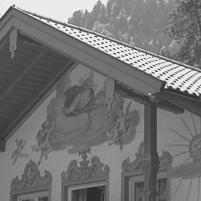}}&
\subfloat[Noisy, blurred]{\includegraphics[height = 62pt]{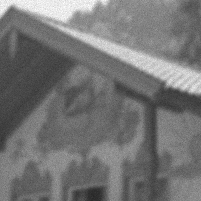}}
\end{tabular}
&
\begin{tabular}{cc}
\subfloat[TV\textsubscript{PSNR}]{\includegraphics[height = 62pt]{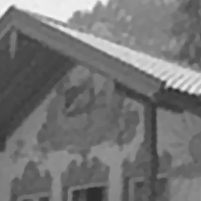}} &
\subfloat[TV\textsubscript{MS-SSIM}]{\includegraphics[height = 62pt]{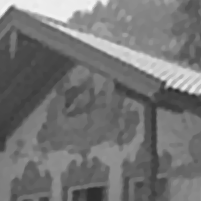}}
\end{tabular}
&
\begin{tabular}{cc}
\subfloat[aTV\textsubscript{PSNR}]{\includegraphics[height = 62pt]{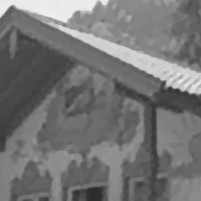}}&
\subfloat[aTV\textsubscript{MS-SSIM}]{\includegraphics[height = 62pt]{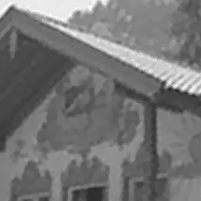}}
\end{tabular}
\end{tabular}
\end{adjustbox}
\caption{TV deblurring and denoising performed with a learned {\em global} (TV\textsubscript{PSNR} and TV\textsubscript{MSSSIM}) and {\em adaptive} (aTV\textsubscript{PSNR} and aTV\textsubscript{MS-SSIM}) regularization hyper-parameter, $p^0$. The writing on the wall (panels a, d) is better restored by the adaptive filters (panels h, i) and it is slightly more contrasted and noisy when MS-SSIM is maximized. The texture of the roof (panels j, m) is better restored by the adaptive filter maximizing MS-SSIM (panel r).}
\label{fig:TVDeblur}
\end{figure*}

The average image quality metrics achieved on the testing dataset (the second half of the Kodak dataset) are reported in Table~\ref{table:TVDeblur}. This shows a slight  advantage of using the adaptive strategy over the non adaptive one, both in the case of the maximization of PSNR and MS-SSIM. Fig.~\ref{fig:TVDeblur}(f) and \ref{fig:TVDeblur}(h) show that the adaptive filter learned by maximizing the PSNR is capable of better reconstructing small details in the image, like the writing on the wall. This is achieved using a larger regularization hyper-parameter close to the edges (Fig.~\ref{fig:TVDeblur}(b) and~\ref{fig:TVDeblur}(k)). Nonetheless, since most of the pixels in the smooth part of the image are still filtered using the global optimum hyper-parameter, the overall gain in terms of PSNR is small. Maximizing MS-SSIM instead of PSNR leads to a slightly different adaptation strategy and to images with more contrast on small details, that are also a bit more noisy (Fig.~\ref{fig:TVDeblur}(g) and~\ref{fig:TVDeblur}(i)).

\begin{table}
\begin{center}
{\small
\begin{tabular}{|c|c|c|c|c|}
\hline
& TV\textsubscript{PSNR} & TV\textsubscript{MS-SSIM} & aTV\textsubscript{PSNR} &  aTV\textsubscript{MS-SSIM}\\
\hline
$p^0$ & 0.0065 & 0.0069 & Eq. \eqref{eq:logistic} & Eq. \eqref{eq:logistic}\\
\hline
PSNR & 28.455 & 28.45 & \bf{28.49} & 28.43\\
SSIM & 0.8745 & 0.8745 & 0.8764 & \bf{0.8774}\\
MS-SSIM & 0.9410 & 0.9411 & 0.9421 & \bf{0.9425}\\
\hline
\end{tabular}
}
\caption{Average image quality metrics on the testing dataset for TV deblurring.~The values of the hyper-regularization parameter $p^0$ are also reported. The best result in each row is shown in bold.}
\label{table:TVDeblur}
\end{center}
\end{table}

\section{Conclusion}
\label{sec:conclusion}

We have presented a general method to learn tuning of parameters in adaptive image processing. We applied it effectively to three classical problems, but the same procedure can be applied to other image processing problems (\eg, segmentation, compression, and so on), once a proper cost function and set of features have been defined.

The proposed learning procedure always led to an improvement in the quality of the processed images. We demonstrated its application in the case of TV deblurring, an iterative processing procedure for which it is not possible to write an analytical cost function to optimize; the improvement in this case is small as a single parameter is optimized and the space of the possible outputs is highly constrained by the filter flow-chart. For a complex filter like ANLM, where two adaptive parameters are considered, we obtained a gain similar to that reported in~\cite{Duv11}, where adaptivness is driven by a statistical risk criterion. For demosaicing, we demonstrated how to apply our method to blend different filters with orthogonal properties; this  has the potential to produce large improvements in the learned adaptive algorithm, and in fact we achieved state-of-the-art results by learning a global mixture, and a further slightly improvement by introducing adaptiveness.

We also observed, consistent with~\cite{Zha15}, that optimizing for different cost functions leads to different parameter modulation strategies. Maximizing for perceptual image quality metrics improves their score and likely also the actual perceived quality. Since we resorted to derivative-free optimization, optimizing for complex, state-of-the-art metrics like FSIM~\cite{Zha11b}, that better correlates with human judgement of image quality, is also doable with our method. We leave these aspects for future investigations.

\bibliography{ref}
\bibliographystyle{IEEEtran}

\end{document}